%% file: root.tex
\documentclass[letterpaper, 10 pt, conference]{ieeeconf}

\IEEEoverridecommandlockouts
\usepackage{graphicx} 
\usepackage{epsfig} 
\usepackage{mathptmx} 
\usepackage{times} 
\usepackage{amsmath} 
\usepackage{amssymb}
\usepackage{siunitx}
\usepackage{xcolor}
\usepackage{caption}                    
\usepackage{subcaption}                 
\usepackage{epstopdf}                   

\usepackage{pgfplots}
\usepgfplotslibrary{groupplots}
\usetikzlibrary{external}
\tikzset{external/system call={lualatex \tikzexternalcheckshellescape -halt-on-error -interaction=batchmode -jobname "\image" "\texsource"}} 
\usepackage[%
xindy,
section = section,
nonumberlist,
sanitize={symbol=false},
shortcuts,
acronym,
nomain,
nowarn
]{glossaries} 

\input{preamble/new_commands}
\input{preamble/glossary}

\title{\LARGE \bf
	Entropic Risk Measure in Policy Search
}

\author{David Nass, Boris Belousov, and Jan Peters
\thanks{The authors are with Intelligent Autonomous Systems Lab,
	Department of Computer Science,
	Technische Universität Darmstadt, Germany,
	{\tt surname@ias.tu-darmstadt.de}}%
\thanks{This project has received funding
from the European Union’s Horizon 2020 research and innovation programme under grant agreement No 640554.}%
}
\begin{document}

\maketitle
\thispagestyle{empty}
\pagestyle{empty}

\begin{abstract}
	With the increasing pace of automation, modern robotic systems need to act in stochastic, non-stationary, partially observable environments. A range of algorithms for finding parameterized policies that optimize for long-term average performance have been proposed in the past. However, the majority of the proposed approaches does not explicitly take into account the variability of the performance metric, which may lead to finding policies that although performing well on average, can perform spectacularly bad in a particular run or over a period of time. To address this shortcoming, we study an approach to policy optimization that explicitly takes into account higher order statistics of the reward function. In this paper, we extend policy gradient methods to include the entropic risk measure in the objective function and evaluate their performance in simulation experiments and on a real-robot task of learning a hitting motion in robot badminton.
\end{abstract}

\input{content/introduction}
\input{content/riskpolicysearch}
\input{content/experiments}
\input{content/conclusion}

\bibliographystyle{IEEEtran} 
\bibliography{IEEEabrv,lit}

\end{document}

%% file: preamble/new_commands.tex
\newcommand*\dif{\mathop{}\!\mathrm{d}}

\DeclareMathOperator{\E}{\mathbb{E}}

\DeclareMathOperator{\N}{\mathcal{N}}

\newcommand{\Var}{\mathrm{Var}}

\newcommand{\T}{{\mathrm{T}}}

\renewcommand{\vec}[1]{\mathbf{#1}}
\newcommand{\vecgrk}[1]{\mbox{\boldmath{$#1$}}}

\newlength\figureheight
\newlength\figurewidth

\definecolor{color0}{rgb}{0.917647058823529,0.917647058823529,0.949019607843137}
\definecolor{color1}{rgb}{0.298039215686275,0.447058823529412,0.690196078431373}
\definecolor{color2}{rgb}{0.333333333333333,0.658823529411765,0.407843137254902}
\definecolor{color3}{rgb}{0.768627450980392,0.305882352941176,0.32156862745098}
\definecolor{color4}{rgb}{0.505882352941176,0.447058823529412,0.698039215686274}
\definecolor{color5}{rgb}{0.8,0.725490196078431,0.454901960784314}
\definecolor{color6}{rgb}{0.392156862745098,0.709803921568627,0.803921568627451}

%% file: preamble/glossary.tex
\newacronym{iid}{i.i.d.}{independently and identically distributed}
\newacronym{RL}{RL}{reinforcement learning}
\newacronym{pg}{PG}{policy gradient}
\newacronym{npg}{NPG}{natural policy gradient}
\newacronym{reps}{REPS}{relative entropy policy search}
\newacronym{rwr}{RWR}{reward weighted regression}
\newacronym{ppo}{PPO}{proximal policy optimization}
\newacronym{mdp}{MDP}{Markov decision process}
\newacronym{dmp}{DMP}{dynamic movement primitive}
\newacronym{promp}{ProMP}{probabilistic movement primitive}
\newacronym{mp}{MP}{movement primitive}
\newacronym{kf}{KF}{Kalman filter}
\newacronym{ekf}{EKF}{extended Kalman filter}
\newacronym{mae}{MAE}{mean absolute error}
\newacronym{mse}{MSE}{mean squared error}
\newacronym{rff}{RFF}{random fourier features}
\newacronym{adam}{Adam}{adaptive moment estimation}
\newacronym{em}{EM}{expectation maximization}
\newacronym{ml}{ML}{maximum likelihood}
\newacronym{crkr}{CrKR}{cost-regularized kernel regression}
\newacronym{trpo}{TRPO}{trust region policy optimization}
\newacronym[\glslongpluralkey={degrees of freedom}]{dof}{DOF}{degree of freedom}
\newacronym[\glsshortpluralkey={KL}]{KL}{KL}{Kullback-Leibler}
\newacronym{LQG}{LQG}{linear quadratic Gaussian}
\newacronym{LEQG}{LEQG}{linear exponential quadratic Gaussian}
\newacronym{CVaR}{CVaR}{conditional value at risk}

%% file: content/introduction.tex
\section{Introduction}
Applying \gls{RL} to robotics is notoriously hard due to the curse of dimensionality~\cite{bellman1957dynamic}.
Robots operate in continuous state-action spaces and visiting every state quickly becomes infeasible.
Therefore, function approximation has become essential to limit the number of parameters that need to be learned.
Policy search methods, that employ pre-structured parameterized policies to deal with continuous action spaces,
have been successfully applied in robotics~\cite{PolSearchSurvey}.
These methods include policy gradient~\cite{williams1992simple,sutton2000policy},
natural policy gradient~\cite{kakade2002npg}, \gls{em} policy search~\cite{kober2009learning,peters2007rwr},
and information theoretic approaches~\cite{peters2010reps}.

A common feature of the aforementioned policy search methods is that they all aim to maximize the expected reward.
Therefore, they do not take into account the variability and uncertainty of the performance measure.
However, robotic systems need to act in stochastic, non-stationary, partially observable environments.
To account for these challenges, the objective function should include an additional variance related criteria.
This paper contributes to the field of reinforcement learning for robotics
by extending the range of applicability of policy search methods to problems with
risk-sensitive optimization criteria, where risk is given by the entropic risk measure~\cite{follmer2011entropic}.

\section{Related work}
Howard and Matheson \cite{howard1972risk} along with Jacobson \cite{jacobson1973LEQG}
were the first to consider risk-sensitivity in optimal control both in discrete and continuous settings.
Jacobson attempted to solve the linear exponential quadratic Gaussian problem that is analogous to the linear quadratic Gaussian
but with an exponentially transformed quadratic cost.
Later, connections between risk-sensitive optimal control and $H_{\infty}$ theory \cite{whittle1990risk}
as well as differential games \cite{fleming1992risk} were found.

In the recent years, there have been some advances in risk-sensitive policy search using policy gradients.
In \cite{tamar2012policy}, a policy gradient algorithm was developed
that accounted for the variance in the objective either through a penalty or as a constraint.
The Conditional value at risk criterion was combined with policy gradients in \cite{prashanth2014policy} and \cite{tamar2015risk}.
In this paper, we study properties of policy gradient methods with the entropic risk measure in the objective.
Employing this particular type of risk measure reveals tight links to popular policy search algorithms,
such as \gls{rwr}~\cite{peters2007rwr} and \gls{reps}~\cite{peters2010reps}.

%% file: content/riskpolicysearch.tex
\section{Background and Notation}

\begin{figure}[b]
	\vspace{-0.5em}
	\centering
	\begin{subfigure}{0.4\linewidth}
		\centering
		\includegraphics[width=\textwidth, trim={12cm 9cm 25cm 3cm},clip]{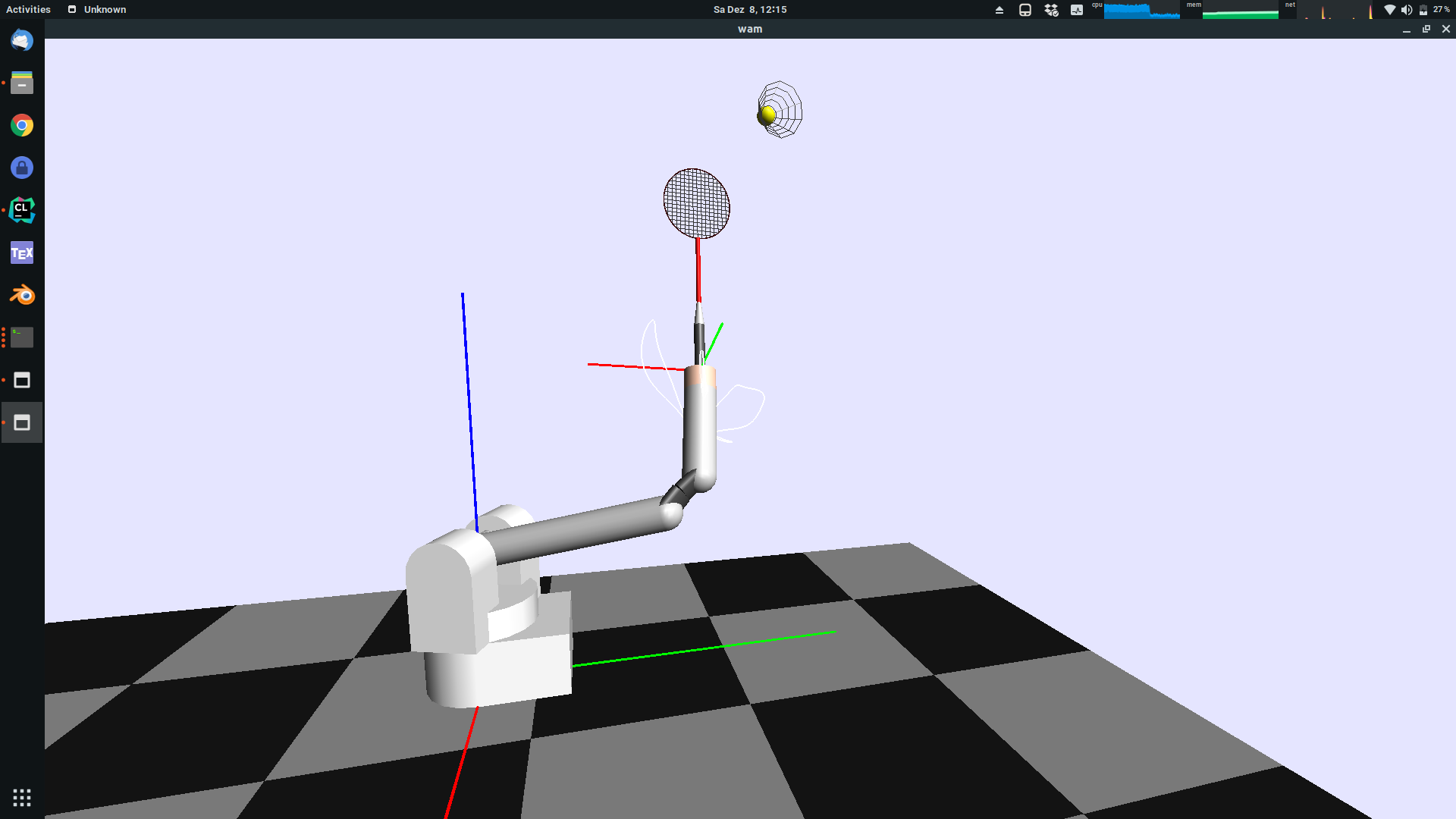}
		\caption{Barrett WAM in a simulation environment}
		\label{fig:SL_badminton}
	\end{subfigure}\quad
	\begin{subfigure}{0.4\linewidth}
		\centering
		\includegraphics[width=\textwidth, trim={10cm 67cm 15cm 14cm},clip]{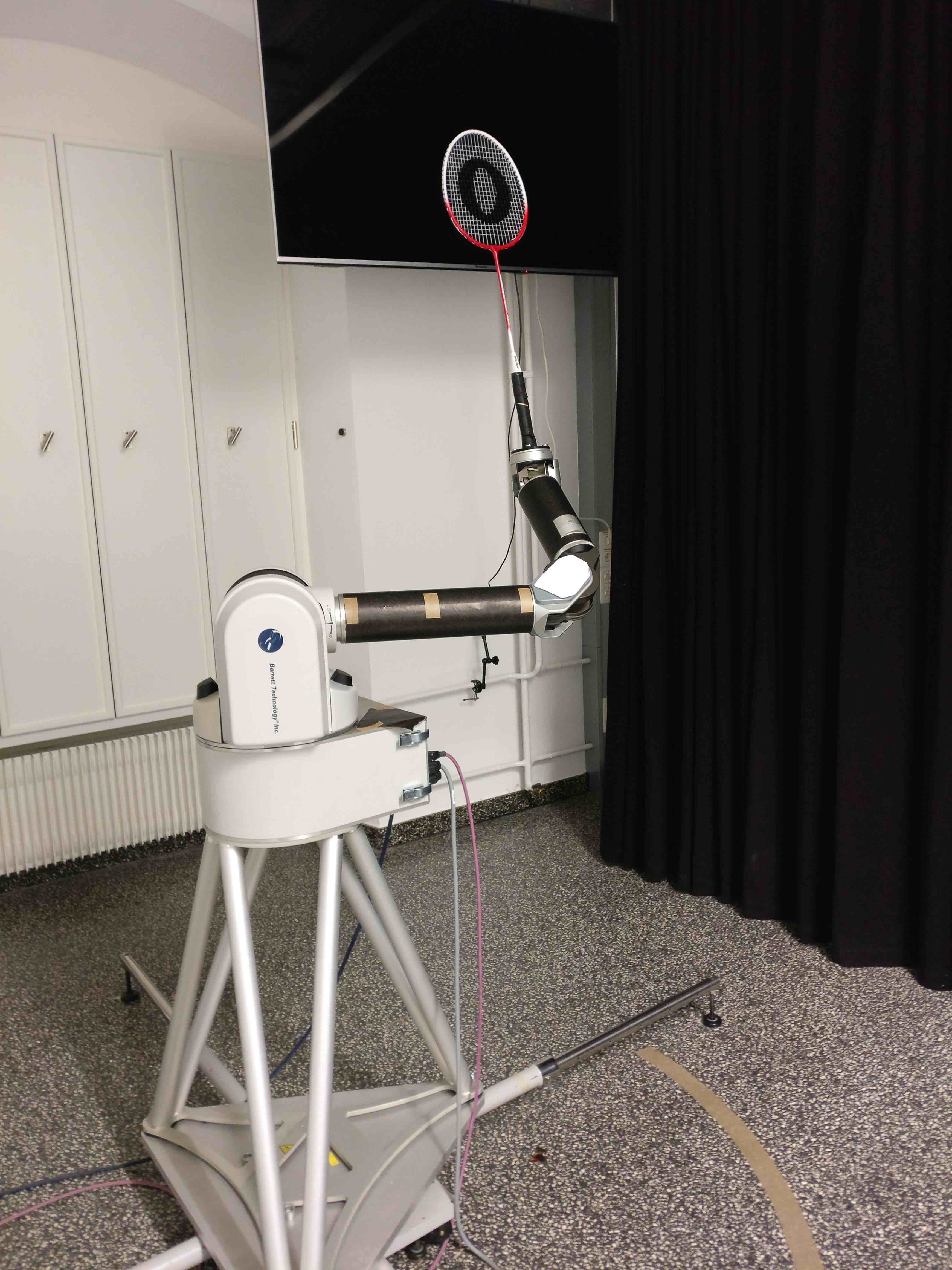}
		\caption{Barrett WAM at IAS, TU Darmstadt}
		\label{fig:barrett_badminton}
	\end{subfigure}
	\caption{Robot badminton evaluation task.}
	\label{fig:setup}
	\vspace{-0.3em}
\end{figure}

In this section, we provide the necessary background information on risk measures and policy search methods.
\subsection{Risk-Sensitive Objectives and Measures} \label{subsec:risk_objectives}

The goal of an \gls{RL} agent is to learn a mapping from states to actions
that maximizes a performance measure~\cite{sutton2018reinforcement}.
In the episodic \gls{RL} setting~\cite{PolSearchSurvey},
the episode return $R$ (also called reward) is a random variable,
and the standard objective is to maximize the expected return.
Our approach considers the case where the performance measure is risk-sensitive.
That is, instead of optimizing for the long-term system performance 
we account for higher order moments of~$R$ by performing a risk-sensitive transformation of the return.

Risk-sensitivity can be described in terms of utility~\cite{whittle2002risk}.
Namely, the utility function defines a transformation of the reward $U(R)$.
The performance measure is then given by the expected utility $\E[U(R)]$.
Clearly, depending on the choice of function $U(R)$, different behaviors will arise.
A special choice of the utility function,
which attracted a lot of interest in various fields~\cite{whittle2002risk},
is given by the exponential function
\begin{equation}\label{key}
	U(R) = \exp(-\gamma R)
\end{equation}
with a risk-sensitivity factor $\gamma \in \mathbb{R}$.
Depending on the sign of $\gamma$, the expected utility based on~\eqref{key} needs to be
either minimized or maximized~\cite{whittle1981risk}, as explained in the following.

For a positive $\gamma > 0$, the expectation $\E[U(R)]$ is a convex decreasing function of $R$,
therefore it needs to be minimized in order to maximize the reward.
In this case, a certain expected reward with lower variance is favored,
and the utility function is called \emph{risk-averse}.
On the other hand, when $\gamma < 0$, the expected utility needs to be maximized,
which leads to favoring high-variance rewards.
In this case the utility is called \emph{risk-seeking}.

To avoid confusion, both cases $\gamma >0$ and $\gamma < 0$ are often treated at once
through the certainty-equivalent expectation~\cite{whittle2002risk}
which always has to be maximized,
\begin{equation} \label{equ:exp_risk_measure}
	J_{\mathrm{risk}}(R) = U^{-1} \E[U(R)] = -\frac{1}{\gamma}\log \E[\exp(-\gamma R)].
\end{equation}
When $\gamma > 0$, this quantity is called the \emph{entropic risk measure}~\cite{follmer2011entropic}.
We slightly abuse terminology and refer to it as the entropic risk measure
in the risk-seeking case $\gamma < 0$ too.

%
\subsection{Policy Search}
\label{subsec:policy_search}

\begin{figure*}[t]
	\centering
	\begin{subfigure}{.3\linewidth}
		\centering
		\includegraphics[width=\linewidth,trim={0.6cm 0.5cm 0.4cm 0.3cm},clip]{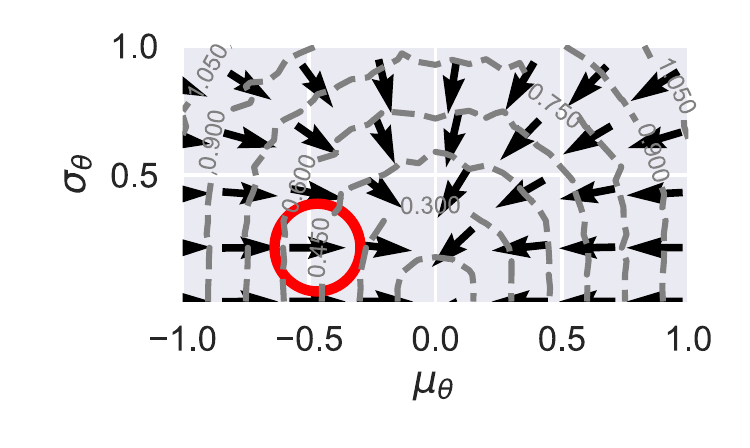}
		\caption{\glstext{pg}, risk-neutral}
	\end{subfigure}%
	\begin{subfigure}{.3\linewidth}
		\centering
		\includegraphics[width=\linewidth,trim={0.6cm 0.5cm 0.4cm 0.3cm},clip]{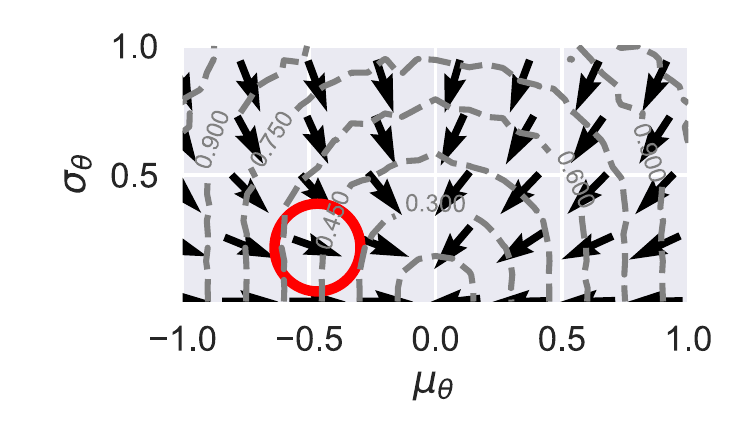}
		\caption{\glstext{pg}, risk-averse}
	\end{subfigure}
	\begin{subfigure}{.3\linewidth}
		\centering
		\includegraphics[width=\linewidth,trim={0.6cm 0.5cm 0.4cm 0.3cm},clip]{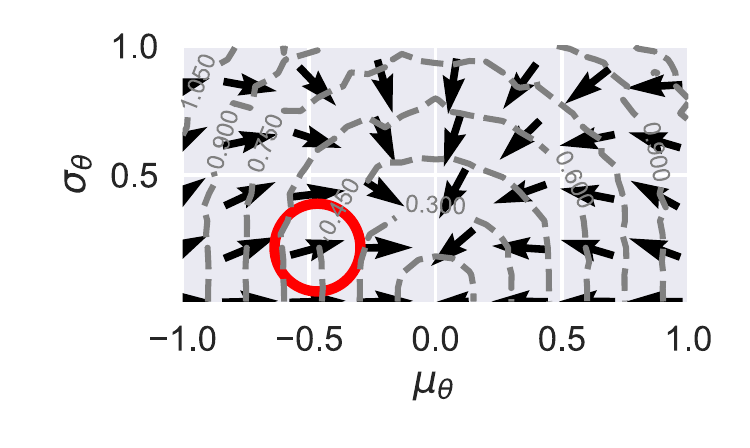}
		\caption{\glstext{pg}, risk-seeking}
	\end{subfigure}%
	\caption{
		Comparison of policy gradient directions on a simple linear system
		with risk-neutral, risk-averse, and risk-seeking objectives.
		Policy mean and variance are parameterized as
		$\mu_\theta = \omega_1$ and $\sigma_\theta = \exp(\omega_2)$.
		Interestingly, risk-seeking \gls{pg} may even point away from the optimum
		if it values uncertainty more than the expected return (see red circle).
	}
	\label{fig:comp_pg_npg}
	\vspace*{-5pt}
	\hrulefill
	\vspace*{-10pt}
\end{figure*}

Consider the finite-horizon episodic \gls{RL} setting~\cite{sutton2018reinforcement}.
At each time step $t$, an agent takes action $\vec{a}_t$ depending on the current state $\vec{s}_t$
by sampling it from a policy $\vecgrk{\pi}_{\vecgrk{\theta}} = \pi(\vec{a}|\vec{s},\vecgrk{\theta})$
parameterized by $\theta$.
Subsequently, the agent transitions into the next state $\vec{s}_{t+1}$
with probability $\mathcal{P}(\vec{s}_{t+1}|\vec{s}_t, \vec{a}_t)$.
Real-valued reward $r(\vec{s}_t, \vec{a}_t)$ is collected at each time-step.
The goal is to find a policy that maximizes the expected return
\begin{equation}
	J_{\vecgrk{\theta}} = \int p_{\vecgrk{\theta}}(\vecgrk{\tau})R(\vecgrk{\tau}) \dif\vec{\tau}
\end{equation} 
where $\vecgrk{\tau} = [\vec{s}_0, \vec{a}_0, \vec{s}_1, \vec{a}_1, ...]$ is a trajectory,
$p_{\vecgrk{\theta}}(\vecgrk{\tau})$ is a distribution over trajectories induced by policy $\vecgrk{\pi}_{\vecgrk{\theta}}$,
and the cumulative reward is defined as $R(\vecgrk{\tau}) = \sum_{t=0}^{T} r(\vec{s}_t, \vec{a}_t)$.

Following common practice of policy search methods~\cite{PolSearchSurvey},
we do not learn parameters $\vecgrk{\theta}$ directly,
but rather learn an upper-level policy $\vecgrk{\pi}_{\vecgrk{\omega}}(\vecgrk{\theta})$
that selects the parameters of the lower-level policy $\vecgrk{\pi}_{\vecgrk{\theta}}$.
Typically, the upper-level policy is modeled as a Gaussian distribution
$\vecgrk{\vecgrk{\theta}}\sim \N(\vecgrk{\mu}_{\vecgrk{\theta}}, \vecgrk{\Sigma}_{\vecgrk{\theta}})$.
By defining the distribution over $\vecgrk{\theta}$, we can explore directly in the parameter space.
The resulting optimization problem for learning the upper-level policy is given by
\begin{align}
	\underset{\vecgrk{\omega}}{\text{maximize}}\quad 
	J({\vecgrk{\omega}}) &= \int \pi_{\vecgrk{\omega}}(\vecgrk{\theta})
		\int p_{\vecgrk{\theta}}(\vecgrk{\tau}) R(\vec{\vecgrk{\tau}}) \dif \vecgrk{\tau} \dif \vecgrk{\theta}
		\label{equ:upper_level_objective} \\
	&= \int \pi_{\vecgrk{\omega}}(\vecgrk{\theta})R(\vec{\vecgrk{\theta}})  \dif \vecgrk{\theta}
	= \E_{\vecgrk{\theta}\sim\pi_{\vecgrk{\omega}}}[R(\vec{\vecgrk{\theta}})].  \nonumber
\end{align}
Mind the common abuse of notation: $R(\theta)$ is not the same function as $R(\tau)$, although they are related.
It is important to note that in the episodic scenario, the low-level policy return~$R(\vecgrk{\theta})$
is not limited to be the cumulative reward function
but can be any function computed over rollouts~\cite{PolSearchSurvey}.

\section{Risk-Sensitive Policy Search}
Expected return~\eqref{equ:upper_level_objective} is the prime optimization objective
for the majority of policy search 
methods~\cite{sutton2000policy,kakade2002npg,peters2007rwr,peters2010reps,kober2012reinforcement}.
To introduce risk-sensitivity into policy search,
we propose to optimize the entropic risk-measure~\eqref{equ:exp_risk_measure} instead.
Rewriting it for the parameters $\vecgrk{\omega}$ of the upper-level policy yields
\begin{equation} \label{equ:risky_policy_search_objective}
 	J_\gamma({\vecgrk{\omega}}) 
	= -\frac{1}{\gamma}\log\E_{\vecgrk{\theta}\sim\pi_{\vecgrk{\omega}}}[\exp(-\gamma R(\vecgrk{\theta}))].
\end{equation}
In the following, policy search methods that maximize the objective~\eqref{equ:risky_policy_search_objective}
are described and studied.
First, a risk-sensitive \gls{pg} algorithm is derived in Sec.~\ref{subsec:risky_policy_gradient}.
After that, a connection to the \gls{reps} algorithm~\cite{peters2010reps}
is established in Sec.~\ref{subsec:risk_reps}.

\subsection{Risk-Sensitive Policy Gradient}
\label{subsec:risky_policy_gradient}
As the name suggests~\cite{PolSearchSurvey},
policy gradient methods aim to maximize the objective $J({\vecgrk{\omega}})$
by gradient ascent on the policy parameters
\begin{equation*}
	\vecgrk{\omega}_{k+1} = \vecgrk{\omega}_k + \alpha \nabla J({\vecgrk{\omega}}_k).
\end{equation*}
The likelihood ratio trick is commonly invoked to derive an estimate of the gradient.
For the risk-sensitive objective~\eqref{equ:risky_policy_search_objective}, the likelihood ratio gradient yields
\begin{equation}
	\nabla J_\gamma
	= \E_{\vecgrk{\theta}\sim\pi_{\vecgrk{\omega}}}
	\left [
		\nabla \log \pi_{\vecgrk{\omega}}(\theta)
		\left\{ -\frac{1}{\gamma} \mathrm{e}^{-\gamma(R(\theta) - \psi_\gamma(\pi_{\vecgrk{\omega}}))} \right\}
	\right]
	\label{equ:gradient_exponential}
\end{equation}
where $\psi_\gamma(\pi_{\vecgrk{\omega}}) = -\gamma^{-1} \log \E_{\pi_{\vecgrk{\omega}}} [\exp(-\gamma R)]$
is the log-partition function~\cite{wainwright2008graphical}.
Expression~\eqref{equ:gradient_exponential} for the risk-sensitive gradient
plays a fundamental role in our further discussion and we will often refer to it in the following.

The first point to make about~\eqref{equ:gradient_exponential}
is the relation between the risk-sensitive policy gradient and the standard, risk-neutral one.
Observe from~\eqref{equ:risky_policy_search_objective} that the risk-sensitive objective
$J_\gamma({\vecgrk{\omega}})$ becomes risk-neutral for $\gamma \to 0$.
That is, by Taylor expansion, one can show that $J_\gamma \to J = \E[R]$.
Surprisingly, however, \emph{the gradient of the risk-sensitive objective does not correspond to the vanilla \gls{pg}}
$\nabla J = \E [\nabla\log \pi  \cdot R]$ but instead to the \gls{pg}
with an average reward baseline
\begin{equation}
	\nabla J_\gamma \xrightarrow[\gamma \to 0]{} \E [\nabla\log \pi  \cdot (R - \E[R])].
	\label{eq:risk_pg_with_baseline}
\end{equation}
The log-partition function $\psi_\gamma(\pi_{\vecgrk{\omega}})$ plays the role of the risk-sensitive baseline,
since $\psi_\gamma \to \E [R]$ for $\gamma \to 0$.
Therefore, risk-sensitive \gls{pg}~\eqref{equ:gradient_exponential}
automatically has lower variance compared to vanilla \gls{pg} due to the presence of the baseline.

Regarding computational aspects,
expectations in~\eqref{equ:gradient_exponential} can be estimated by averages
$\E_{\vecgrk{\theta}\sim\pi_{\vecgrk{\omega}}} [f(\vec{\vecgrk{\theta}})]
= N^{-1}\sum_{i=1}^{N} f(\vec{\vecgrk{\theta_i}})$.
Furthermore, we can view~\eqref{equ:gradient_exponential}
as a risk-neutral \gls{pg} for an exponentially transformed reward function
given by the expression in curly braces in~\eqref{equ:gradient_exponential}.
Therefore, along with the multiplicative baseline $\psi_\gamma(\vecgrk{\omega})$,
the usual additive baseline can also be subtracted to further reduce variance.
Moreover, standard algorithms,
such as \gls{npg}~\cite{kakade2002npg} and \gls{ppo}~\cite{schulman2017ppo},
can be directly applied to optimize the risk-sensitive objective~\eqref{equ:risky_policy_search_objective}
thanks to the form~\eqref{equ:gradient_exponential} of the risk-sensitive policy gradient.

\subsection{Parameter Exploration with Risk-Sensitivity}
Exploration noise and reward variability may come in conflict within the risk-sensitive optimization framework.
In episodic policy search~\cite{PolSearchSurvey},
exploration is achieved by sampling parameters $\vecgrk{\theta}$ of a lower-level policy $\pi_{\vecgrk{\theta}}$
from a stochastic upper-level policy $\pi_{\vecgrk{\omega}} = \N(\mu_\theta, \sigma_\theta^2)$.
Variance of the upper-level policy determines the granularity at which parameter space is queried.
At the same time, it directly affects variability of the observed rewards.
Therefore, reward variability gets entangled with exploration noise.

We proceed to examine the following one-dimensional toy problem
\begin{equation}
	\underset{\mu_\theta, \sigma_\theta^2}{\text{maximize}}
	\quad -\frac{1}{\gamma}\log\E_{\vecgrk{\theta}\sim \N\left(\mu_\theta, \sigma_\theta^2\right)}
		\left[ \mathrm{e}^{\gamma \left| \vecgrk{\theta} \right|} \right].
	\label{equ:lin_ex}
\end{equation}
Policy gradients of the objective~\eqref{equ:lin_ex},
evaluated on a grid of parameter values $\mu_\theta$ and $\sigma_\theta$,
are displayed in Fig.~\ref{fig:comp_pg_npg}.
A risk-neutral ($\gamma = 0$), risk-averse ($\gamma > 0$), and risk-seeking ($\gamma < 0$) scenarios are shown.
Covariant parameterization~\cite{wierstra2008natural} of the Gaussian density was used,
i.e., $\mu_\theta = \omega_1$ and $\sigma_\theta = \exp(\omega_2)$;
therefore, the displayed gradient directions coincide with the natural gradient directions for this problem.
A crucial observation from Fig.~\ref{fig:comp_pg_npg}
is that a risk-seeking policy update may increase exploration variance,
whereas a risk-averse update always decreases it.
In practical terms, risk-aversion may lead to premature convergence due to insufficient exploration.
Optimism, on the other hand, may result in a better coverage of the search space by fostering exploration.

\subsection{Connection to Relative Entropy Policy Search} \label{subsec:risk_reps}
In Sec.~\ref{subsec:risky_policy_gradient}, we established a remarkable fact that
risk-sensitive \gls{pg}~\eqref{equ:gradient_exponential}
yields a baseline-corrected gradient estimator~\eqref{eq:risk_pg_with_baseline}
in the risk-neutral limit $\gamma \to 0$.
It turns out, another important property of the gradient estimator~\eqref{equ:gradient_exponential}
can be revealed by recognizing it as the gradient
of the \gls{ml} policy update in \Gls{reps}~\cite{peters2010reps}.
This renders our risk-sensitive policy update optimal in a certain information-theoretic sense, made precise below.

\Gls{reps} belongs to the category of information-theoretic policy search approaches~\cite{PolSearchSurvey}.
This class of methods follows the idea of limiting the loss of information in-between policy updates.
The \gls{KL} divergence is commonly used as the measure of information loss.
\Gls{reps} can be framed as an \gls{em}-like algorithm,
with the parameter update step given by the weighted \gls{ml} fit \cite{peters2010reps}.
At each iteration, the following optimization problem gets solved
\begin{equation}
\begin{aligned}
	\underset{\vecgrk{\pi}}{\mathrm{maximize}}
		&\quad \int \vecgrk{\pi}(\vecgrk{\theta})R(\vec{\vecgrk{\theta}}) \dif \vecgrk{\theta} \\
	\text{subject to}
		&\quad  {\mathrm{KL}}(\vecgrk{\pi}(\vecgrk{\theta}) \| q(\vecgrk{\theta})) \le \epsilon, \\
		&\quad \int \vecgrk{\pi}(\vecgrk{\theta})\dif \vecgrk{\theta} = 1.
	\label{equ:reps_optim}
\end{aligned}
\end{equation}
Conveniently, a closed-form solution can be found
\begin{equation} \label{equ:reps_closed_form}
	\pi(\vecgrk{\theta}) = q(\vecgrk{\theta}) \exp \left( \frac{R(\vecgrk{\theta})-\psi_{-1/\eta}(q)}{\eta}  \right)
\end{equation}
as a function of the Lagrange multiplier $\eta > 0$,
which corresponds to the \gls{KL}-bound in~\eqref{equ:reps_optim}.
Note the appearance of the log-partition function again, $\psi_{-1/\eta}(q) = \eta \log \E_q [\exp(R/\eta)]$.
The optimal value of $\eta$ is found by dual optimization
\begin{equation} \label{equ:reps_dual}
	\eta^\star = \arg \min_{\eta > 0} \left\{  \eta \epsilon + \psi_{-1/\eta}(q) \right\}.
\end{equation}
Since only black box access to function $R(\vecgrk{\theta})$ is assumed,
Eq.~\eqref{equ:reps_closed_form} does not yield the new policy $\pi$
as an explicit function of $\vecgrk{\theta}$ but rather only provides samples from it.

A parametric policy $\vecgrk{\pi}_{\vecgrk{\omega}}(\vecgrk{\theta})$
is fitted to the samples obtained from \eqref{equ:reps_closed_form} by moment projection~\cite{PolSearchSurvey}
\begin{alignat}{3} 
	\underset{\vecgrk{\omega}}{\mathrm{minimize}}
		&\;\; \mathrm{KL}(\vecgrk{\pi}(\vecgrk{\theta}) \| \vecgrk{\pi}_{\vecgrk{\omega}}(\vecgrk{\theta})) 
		\label{equ:reps_fit_pol} \\
	\propto \underset{\vecgrk{\omega}}{\mathrm{maximize}}
		&\;\; \E_{\vecgrk{\theta} \sim q} \left[ \log\vecgrk{\pi}_{\vecgrk{\omega}}(\vecgrk{\theta}) 
			\exp \left( \frac{R(\vecgrk{\theta}) - \psi_{-1/\eta}(q)}{\eta} \right)\right].
		\nonumber
\end{alignat}
The gradient of~\eqref{equ:reps_fit_pol} serves as the link
to the risk-sensitive policy gradient~\eqref{equ:gradient_exponential}.
Indeed, compare
\begin{equation}
	\nabla_{\vecgrk{\omega}} \mathrm{KL}
	= \E_{\vecgrk{\theta} \sim q}
	\left[
		\nabla \log\vecgrk{\pi}_{\vecgrk{\omega}}(\vecgrk{\theta})
		\left\{ \mathrm{e}^{\eta^{-1} (R(\theta) - \psi_{-1/\eta}(q))} \right\}
	\right]
	\label{equ:grad_reps}
\end{equation}
to the risk-sensitive gradient~\eqref{equ:gradient_exponential}.
The correspondence is established by identifying $\gamma = -1/\eta$
and noting that the argument in the curly braces in~\eqref{equ:grad_reps}
is proportional to the one in~\eqref{equ:gradient_exponential} up to a scaling factor $\eta$.

The key difference between~\eqref{equ:gradient_exponential} and~\eqref{equ:grad_reps}
is the sampling distribution.
Whereas the risk-sensitive policy gradient~\eqref{equ:gradient_exponential}
requires samples from~$\vecgrk{\pi}_{\vecgrk{\omega}}$,
an auxiliary distribution $q$ is used in \gls{reps}.
In theory, it means that \gls{reps} can perform several gradient update steps
according to~\eqref{equ:grad_reps} with the same samples from $q$,
while the risk-sensitive gradient~\eqref{equ:gradient_exponential}
requires gathering new data after each parameter update.
However, in practice, optimizing the \gls{ml} objective~\eqref{equ:reps_fit_pol}
till convergence is problematic due to the finite sample size
and the associated overfitting problems~\cite{PolSearchSurvey}.
That is why alternatives to the policy update objective~\eqref{equ:reps_fit_pol}
are often used, such as performing the information projection
instead of the moment projection~\cite{neumann2011variational},
or constraining the policy fitting step with another \gls{KL} divergence~\cite{abdolmaleki2017deriving},
which can be viewed as a form of maximum a posteriori estimation~\cite{abdolmaleki2018maximum}.

Thus, the policy update of \gls{reps}~\eqref{equ:reps_fit_pol}
can be identified with the risk-sensitive update~\eqref{equ:gradient_exponential}
under the assumption that the information loss bound $\epsilon$ is small,
such that $q \approx \vecgrk{\pi}_{\vecgrk{\omega}}$
and one step in the direction of the gradient~\eqref{equ:grad_reps} solves~\eqref{equ:reps_fit_pol}.
Importantly, though, the temperature parameter $\eta = -1/\gamma$ gets optimized in \gls{reps}
and thus changes with iterations,
whereas when applying~\eqref{equ:gradient_exponential}, it has to be scheduled manually.

Another interesting distinction between risk-sensitive optimization and \gls{reps}
stems from the fact that the temperature parameter $\eta$ must be positive in \gls{reps}.
This means $\gamma < 0$, or risk-seeking optimization.
Thus, \gls{reps} is risk-seeking by construction,
unlike the risk-sensitive \gls{pg}~\eqref{equ:gradient_exponential} which can also be risk-averse.

%% file: content/experiments.tex
\section{Experiments}
\label{sec:experiments}

To analyze the properties of the risk-sensitive policy gradient algorithm of Sec.~\ref{subsec:risky_policy_gradient},
we first consider a prototypical risk-sensitive portfolio optimization problem to establish the validity of our approach,
then we proceed to apply the risk-sensitive policy gradient method
to a toy dynamical system that models a part of our robot badminton setup,
and finally, we report the results obtained by applying the algorithm
to a real-robot task of learning to return a shuttlecock in the game of badminton with the Barrett WAM robot.

\subsection{Risk-Sensitive Portfolio Management}
A basic problem of portfolio optimization can be described as follows~\cite{boyd2017multi}.
An individual wants to invest a unit of capital in $N$ assets with the goal of making profit.
The distribution of capital over assets $\vec{x}$ is called portfolio;
by definition, portfolio is normalized, $\sum_{i=1}^{N}x_i = 1$.
The returns of various assets~$\vec{r}$ are random variables and are assumed to be Gaussian distributed 
$\vec{r} \sim \N(\vecgrk{\mu}_\vec{r},\vecgrk{\Sigma}_\vec{r})$.
Then, return of a portfolio $\vec{x}$ is a random variable
$R \sim \N(\vecgrk{\mu}_\vec{r}^\T\vec{x},\vec{x}^\T\vecgrk{\Sigma}_\vec{r}\vec{x})$.
Depending on the definition of `making profit',
different objective functions can be constructed.
We explore the notion of optimality with respect
to the exponential risk measure~\eqref{equ:exp_risk_measure},
which allows for controlling the mean-variance trade-off (and higher moments)
by varying the risk-sensitivity factor $\gamma$.

To apply the risk-sensitive policy gradient approach from Sec.~\ref{subsec:risky_policy_gradient},
a suitable policy must be defined.
We let the lower-level policy output a portfolio $\vec{x}$,
parameterized by the softmax $\vec{x} = \exp (\vec{\theta} - \log (1^T \exp(\theta)))$.
Parameters $\vecgrk{\theta}\in \mathbb{R}^{N}$ of the lower-level policy
are sampled from a Gaussian upper-level policy
$
	\vecgrk{\pi}_{\vecgrk{\omega}} (\vecgrk{\theta})
	= \N(\vecgrk{\theta}|\vecgrk{\mu}_{\vecgrk{\theta}}, \vecgrk{\Sigma}_{\vecgrk{\theta}})
$
with $\vecgrk{\omega} = \{\vecgrk{\mu}_{\vecgrk{\theta}}, \vecgrk{\Sigma}_{\vecgrk{\theta}}\}$.

In simulation, the number of assets is set to $N = 30$.
Parameters of the asset return distribution are sampled evenly in the interval
$\vecgrk{\mu}_\vec{r}\in [4, 0.5]$, $\vecgrk{\sigma}_\vec{r} \in [2, 0.01]$,
with $\vecgrk{\Sigma}_\vec{r}=\mathrm{diag}(\vecgrk{\sigma}_\vec{r}^2)$.
This distribution of parameters can be interpreted as follows.
Returns with a high expected value are accompanied with higher risks,
whereas lower risk returns yield lower mean reward.
When comparing two policies $\vecgrk{\pi}_1$ and $\vecgrk{\pi}_2$
corresponding to risk factors $\gamma_1 > \gamma_2$,
policy $\vecgrk{\pi}_1$ will prefer lower risk assets and yield lower return on average than $\vecgrk{\pi}_2$.

\begin{figure}[t]
	\vspace{0.5em}
	\centering
	\setlength\figureheight{0.6\linewidth}
	\setlength\figurewidth{\linewidth}
	\resizebox{0.8\linewidth}{!}{
	\hspace{-2em}
	\includegraphics{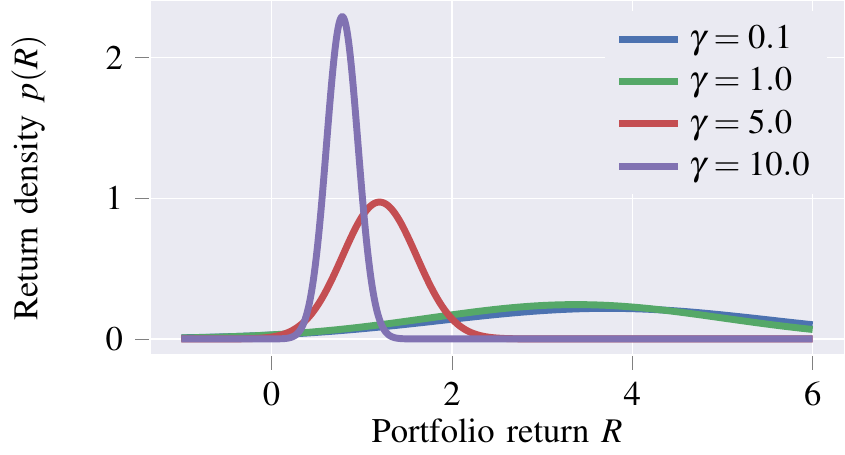}}
	\caption{
		Return distributions of optimal portfolios
		found with risk-sensitive policy gradient~\eqref{equ:gradient_exponential}
		for \mbox{$\gamma \in \{0.1, 1, 5, 10\}$}.
		The higher the risk-aversion factor $\gamma > 0$, the lower the variance of the returns;
		however, the mean is also lower in this case.
	}
	\label{fig:finance_vpg_rew_dist}
	\vspace{-1em}
\end{figure}

Return distributions for various values of the risk-aversion factor $\gamma$
are shown in Fir.~\ref{fig:finance_vpg_rew_dist}.
Simulation was run over $10$ random seeds with $1000$ samples per trial.
Results confirm the theory.
A more pessimistic objective (higher $\gamma$) leads to a narrow reward distribution with lower mean,
implying a smaller but more consistent reward.
Smaller values of $\gamma$, and therefore more risk-neutral objectives, on the other hand,
lead to an asset distribution that almost entirely aims to maximize the expected reward
not taking the variance into account.

\subsection{Toy Badminton System}

We consider a simplified scenario of a robot learning to return a shuttlecock in the game of badminton.
We assume a two dimensional world and a ball following a parabolic flight trajectory.
The goal is to determine the hitting velocity of the racket which results in the ball arriving at a desired target location.
The hypothesis is that for different values of the risk-aversion factor~$\gamma$,
the agent will learn different strategies: either aggressive hits but with high variability,
or safe returns however with smaller expected reward.
The problem is specified as follows
\begin{alignat}{4}
	&\underset{\vecgrk{\omega}}{\text{minimize}}
		\quad &&\frac{1}{\gamma}\log \E \left [ \exp(\gamma|x_\mathrm{des} - x_1)|) \right ] 
	\label{equ:obj_toy_badminton}\\
	&\text{subject to}
		\quad &&x_1 = x_0+v_{x,0} \left ( \frac{v_{y,0}}{g} 
			+ \sqrt{\frac{v_{y,0}^2}{g^2}+\frac{2y_0}{g}} \right ). \nonumber
\end{alignat}
The initial position of the ball $(x_0, y_0)$ is assumed known.
Due to a perfectly elastic collision, the initial velocity of the ball equals the velocity of the racket.
Therefore, we treat the initial ball velocity as the control variable.
Keeping in mind that this model should resemble the real-robot setup considered later,
we add a bit of noise to the initial ball velocity,
such that $(v_{x,0}~v_{y,0}) = \vec{v}_{0} \sim \N(\vec{u},\vecgrk{\Sigma_{\vec{v}_0}})$
with $\vec{u}$ being our control variable.
Constant $g$ is the gravitational constant,
and the equality constraint in~\eqref{equ:obj_toy_badminton}
is derived from the equations of motion.
Optimization variable $\vecgrk{\omega} = \{\vecgrk{\mu}_\vec{u}, \vecgrk{\Sigma}_\vec{u}\}$
contains the parameters of the higher-level policy.
As usually, we employ a Gaussian policy
$\vec{u}\sim\vecgrk{\pi}_{\vecgrk{\omega}}(\vec{u}) = \N(\vec{u}|\vecgrk{\mu}_{\vec{u}}, \vecgrk{\Sigma}_{\vec{u}})$.

\begin{figure}[t]
	\centering
	\begin{subfigure}{.49\linewidth}
		\centering
		\setlength\figureheight{0.6\linewidth}
		\setlength\figurewidth{1.2\linewidth}
		\resizebox{!}{1.4\linewidth}{
			\includegraphics{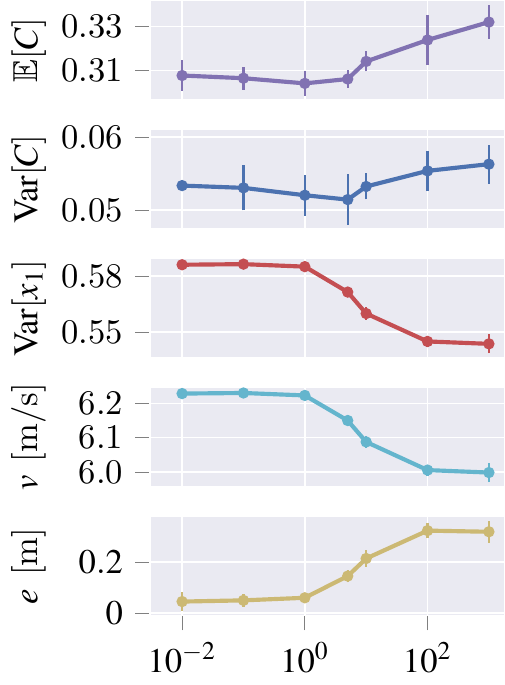}}
		\caption{risk-averse, $\gamma>0$}
		\label{subfig:toy_npg_ra}
	\end{subfigure}%
	\hspace{0.1em}
	\begin{subfigure}{.49\linewidth}
		\centering
		\setlength\figureheight{0.6\linewidth}
		\setlength\figurewidth{1.2\linewidth}
		\resizebox{!}{1.4\linewidth}{
			\includegraphics{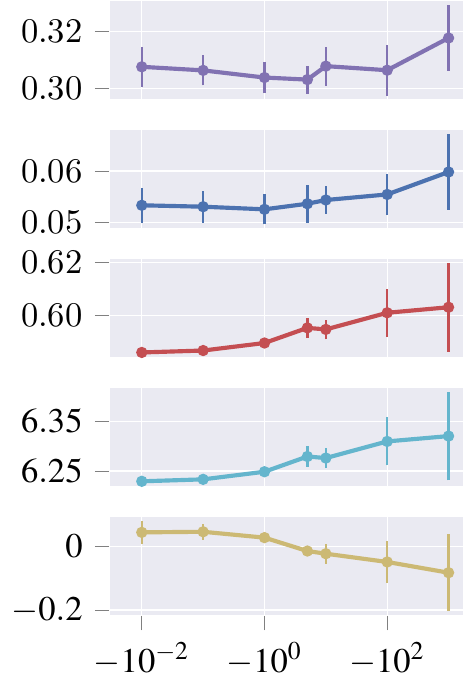}}
		\caption{risk-seeking, $\gamma<0$}
		\label{subfig:toy_npg_rs}
	\end{subfigure}	
	\caption{Simulation results on the toy badminton system 
		with a risk-sensitive objective optimized
		for varying risk factor values $\gamma\in\pm\{0.01, 0.1, 1, 5, 10, 100, 1000\}$.
		System noises are fixed $\sigma_{v_{x,0}}=\sigma_{v_{x,0}} = 0.6$
		and the initial position is $x_0=y_0=0$.} 
	\label{fig:toy_npg}
	\vspace{-1em}
\end{figure}

Evaluation results on this simulated problem are shown in Fig.~\ref{fig:toy_npg}.
Optimization was run over a range of values of $\gamma$
with $1000$ samples per trial, averaged over $10$ random seeds.
Error $e$ is defined as \mbox{$e=x_\mathrm{des} - x_1$}, the cost is $C = |e|$ and $v$ is the initial speed.
Several trends can be observed in Fig.~\ref{fig:toy_npg}.
First, variance in the landing location~$x_1$ is inversely proportional to risk-aversion:
when risk-aversion increases, variance in the landing location decreases.
However, such a clear trend cannot be observed in the variance of the cost function.
Second, from the plot of the final position error $e$, we can read that both risk-seeking and risk-averse
policies corresponding to extreme values of~$\gamma$ fail at returning the ball to the desired target.
This effect is due to the dual nature of the objective function which trades mean performance agains variability.
Extreme risk-averse policies tend to undershoot the target, while extreme risk-seeking ones tend to overshoot it.
The same conclusion can be made based on the plot of velocities~$v$.
Risk-averse, pessimistic policies favor smaller initial velocities.
In contrast, risk-seeking policies chose larger initial velocities.
Third, variance bars are larger for large negative values of~$\gamma$.
This effect is due to objective~\eqref{equ:obj_toy_badminton} becoming very sharp,
close to a delta function, which negatively affects optimization.

\subsection{Robot Badminton}

Finally, we proceed to apply the risk-sensitive policy gradient
on a real robotic system consisting of a Barrett WAM
supplied with an optical tracking setup and equipped with a badminton racket (see Fig.~\ref{fig:setup}).
The goal is to learn movement primitives of different levels of riskiness:
on the scale between an aggressive smash and a defensive backhand.

To represent movements, we encode them using
probabilistic movement primitives (ProMPs)~\cite{ProMPparaschos2013}.
Since a ProMP is given by a distribution over trajectories,
generalization is accomplished through probabilistic conditioning,
and trajectories can be shaped as desired by including via-points.
We utilize these properties of ProMPs to learn hitting movements
encoded by a small set of meta-parameters~\cite{kober2012reinforcement}.

\begin{figure}[t]
	\centering
	\begin{subfigure}{.25\linewidth}
		\centering
		\includegraphics[width=\textwidth]{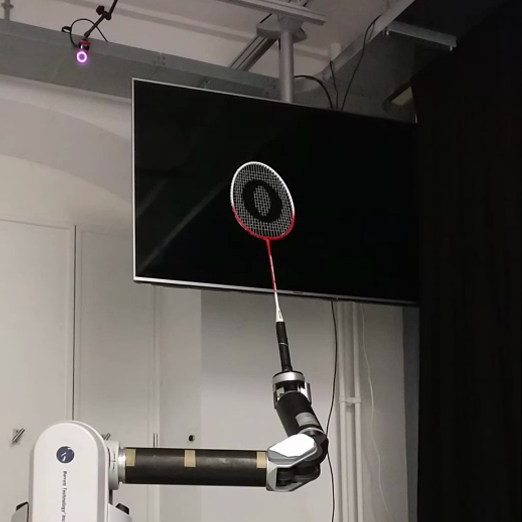}
		\caption{rest}
		\label{subfig:rest}	
	\end{subfigure}%
	\begin{subfigure}{.25\linewidth}
		\centering
		\includegraphics[width=\textwidth]{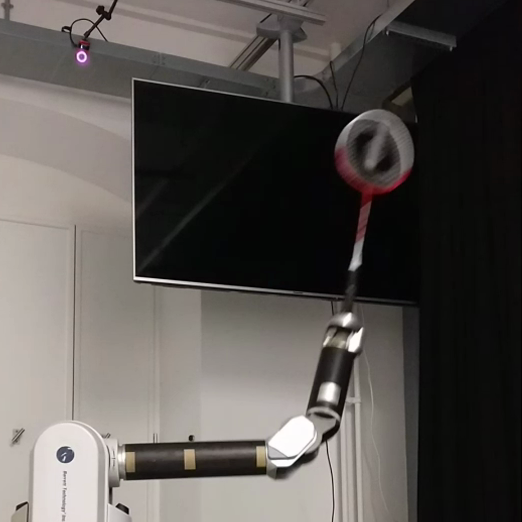}
		\caption{hit}
		\label{subfig:hit}	
	\end{subfigure}%
	\begin{subfigure}{.25\linewidth}
		\centering
		\includegraphics[width=\textwidth]{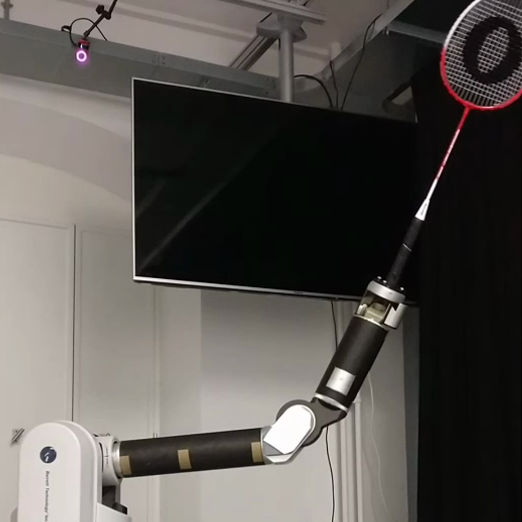}
		\caption{swing}
		\label{subfig:swing}	
	\end{subfigure}%
	\begin{subfigure}{.25\linewidth}
		\centering
		\includegraphics[width=\textwidth]{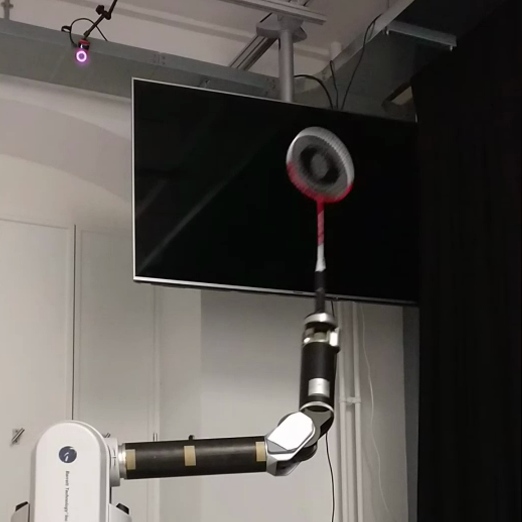}
		\caption{retract}
		\label{subfig:retract}	
	\end{subfigure}%
	\caption{Phases of the hitting movement of the Barrett WAM.}
	\label{fig:hitting squence}	
	\vspace{-1em}
\end{figure}


\begin{figure}[b]
	\vspace{-1em}
	\centering
	\setlength\figureheight{0.5\linewidth}
	\setlength\figurewidth{0.75\linewidth}
	\resizebox{\linewidth}{!}{
		\includegraphics{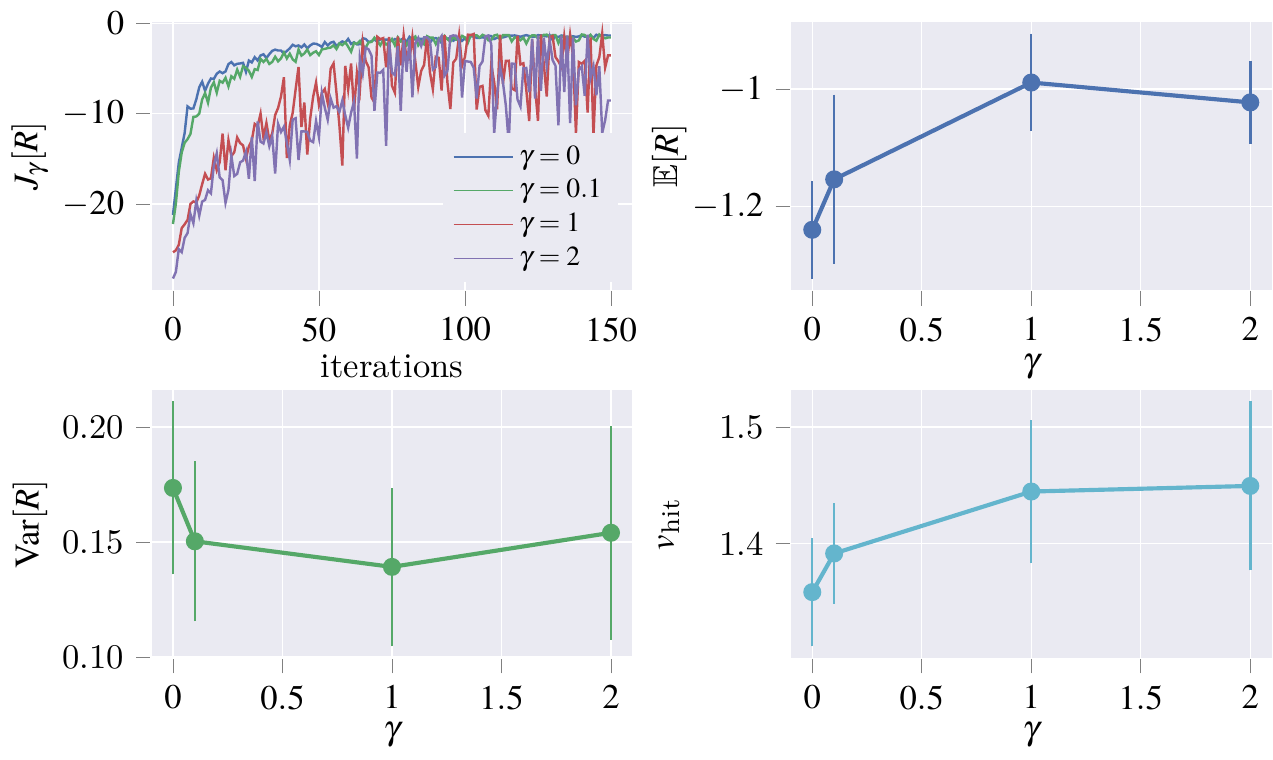}}
	\caption{
	The upper left plot shows the convergence curves for different values of~$\gamma \geq 0$.
	The other three plots compare performance of the final learned policies
	in terms of expected return $\E[R]$, variance $\Var[R]$, and hitting velocity $v_{\text{hit}}$.
	}
	\label{fig:wam_ppo}
	\vspace{-0.4em}	
\end{figure}

The meta parameters in our case encode a via-point
defined by joint positions and velocities of the robot arm at a desired hitting time,
$\vecgrk{\theta} = [\vec{q}_{t,\mathrm{hit}}^\T \quad \dot{\vec{q}}_{t,\mathrm{hit}}^\T]^\T$.
Using the \gls{ekf}, we process shuttlecock observations and predict its future state
\mbox{ $\vec{s} = [\vec{x}_b^\T \quad \dot{\vec{x}}_b^\T]^\T$} at the interception plane.
Control policy
\mbox{$\pi_{\vecgrk{\omega}}(\vecgrk{\theta}|\vec{s})
= \N(\vecgrk{\theta} | \vec{M}^\T\vecgrk{\phi}(\vec{s}), \vecgrk{\Sigma}_{\vecgrk{\theta}})$}
maps state $\vec{s}$ to meta-parameters~$\vecgrk{\theta}$, where
random Fourier features~$\vecgrk{\phi}(\vec{s})$ are tuned 
as described in~\cite{rajeswaran2017sinfeat}. 
Reward function~$R = -\sum_{i\in{x,y,z}}|x_{i,\mathrm{ball}} - x_{i,\mathrm{racket}}| - r_{\mathrm{target}}$ is based on two terms:
it encourages contact between the racket and the shuttlecock
and it provides a bonus $r_{\mathrm{target}}$ if the shuttlecock reaches the desired target afterwards.

This problem falls into the realm of contextual policy search~\cite{PolSearchSurvey}.
Therefore, we state the optimization objective as follows
\begin{equation}\label{equ:badminton_objective}
\underset{\vecgrk{\omega}}{\mathrm{maximize}} \quad
	-\frac{1}{\gamma} \log \int \mu(\vec{s}) \int \pi_{\vecgrk{\omega}}(\vecgrk{\theta}|\vec{s})
	e^{-\gamma R(\theta,\vec{s})} \dif \vecgrk{\theta}  \dif \vec{s},
\end{equation}
where $\mu(\vec{s})$ is the distribution over contexts $\vec{s}$,
and $R(\vecgrk{\theta}, \vec{s})$ is the reward given for the combination of $\vec{s}$ and $\vecgrk{\theta}$.

Returning a shuttlecock in badminton to a desired location requires a high degree of precision. 
In our experiments, we had to relax the requirements due to constraints of the hardware. 
We only optimized for returning the shuttlecock at all and forced the projectile to follow the same trajectory for every iteration and always hit the same point at the interception plane.

Evaluation results are shown in Fig.~\ref{fig:wam_ppo}.
Policies are optimized in simulation using $35$ roll-outs per iteration over $150$ iterations
and experiments are repeated over $5$ random seeds.
Only risk-sensitive policies are shown, but it is noted that risk-seeking also yielded converging results.
Although the convergence plots for positive values of~$\gamma$ look noisier (upper left plot),
the final performance achieved by the policies trained with non-zero risk aversion
(e.g., $\gamma = 1$ or $\gamma = 2$) is higher in terms of the expected reward (upper right plot).
It is hard to make a conclusive judgement about the variance (lower left plot)
due to high variability in the results.
Surprisingly, the hitting velocity~$v_{\text{hit}}$ is actually higher for more risk-averse policies.
This observation stands in contrast to what we had in the toy badminton model,
where less aggressive policies favored smaller velocities.

Unfortunately, with our current setup,
we were not able to achieve the goal of training skills of varying degree of riskiness. 
Nevertheless, we wanted to test the limits of achievable performance in the badminton task following a risk-neutral objective.
The interception of the shuttlecock and hitting plane was enlarged to cover an area of approximately $1\si{\meter^2}$.
We carried out an extended learning trial in simulation with $100$ roll-outs per iteration over $800$ iterations.
The best risk-neutral controller could return $95\%$ of the served balls. The learned policy could be transferred to the real robot and was able to successfully return a shuttle cock. An example hitting movement is shown in Fig.~\ref{fig:hitting squence}.

%% file: content/conclusion.tex
\section{Conclusion}
The entropic risk measure was considered as the optimization objective for policy gradient methods.
By analyzing the exact form of its gradient,
we found that it is related to the standard policy gradient but inherently incorporates a baseline.
Furthermore, risk-sensitive policy update was shown to correspond
to a certain limiting case of the policy update in \gls{reps}.
Exploring this connection to information-theoretic methods appears to be a fruitful direction for future work.
Entanglement between exploration variance and inherent system variability was found to be a strong
limiting factor. Approaches for separating these two sources of uncertainty need to be searched for.

To reveal strengths and weaknesses of risk-sensitive optimization in a real robotic context,
we applied our policy gradient method to the problem of learning risk-sensitive movement primitives
in a badminton task. In a simplified model, we observed that policies optimized for different values of risk aversion
demonstrate qualitatively different behaviors. Namely, risk-averse policies hit the shuttlecock with smaller velocity
and tended to undershoot, whereas risk-seeking policies favored higher velocities and typically overshot the target.
Finally, we carried out experiments on the real robot, which showed that moderate values of risk aversion
can help finding better solutions for the original, risk-neutral problem.
However, our attempt at learning risk-sensitive movement primitives on the real robot had limited success
due to limitations of the hardware platform and the entanglement of sources of variability.